\documentclass[11pt]{article}

\usepackage[preprint]{acl}
\usepackage{amssymb} 
\usepackage{amsmath} 
\usepackage{times}
\usepackage{latexsym}

\usepackage[T1]{fontenc}

\usepackage[utf8]{inputenc}

\usepackage{microtype}

\usepackage{inconsolata}

\usepackage{graphicx}

\usepackage{graphicx}
\usepackage{subcaption}
\usepackage{amsmath}
\usepackage{booktabs}
\usepackage{arydshln}

\title{On the Position Bias of Large Language Models: A Comprehensive Study}

\author{
Yining Li$^{\heartsuit}$\thanks{Correspondence to Yining Li (\texttt{\url{liyn9923@mails.jlu.edu.cn}}) and Huayang Li (\texttt{\url{li.huayang.lh6@is.naist.jp}}).}\ \ \ \ \ Huayang Li$^{\clubsuit}$\ \ \ \ \ Lemao Liu$^{\spadesuit}$ \ \ \ \ \ Taro Watanabe$^\clubsuit$
  \\
  \\
   $^\heartsuit$Jilin University\ \ \ \ \ $^\spadesuit$Nara Institute of Science and Technology\ \ \ \ \
   $^\clubsuit$Fudan University\\
  \\
}

\begin{document}
\maketitle
\begin{abstract}
Recent studies have shown that position bias remains a critical vulnerability in large language models (LLMs), undermining both the integrity of multiple-choice evaluations and the reliability of retrieval-augmented generation.
However, existing understanding of this phenomenon is largely derived from studies with limited model coverage and English-centric benchmarks, leaving the behavior of frontier models and diverse languages underexplored.
In this work, we conduct a comprehensive investigation of position bias across model scales ranging from billion-level to trillion-level parameters, covering both multiple-choice and multi-document tasks in five languages.
Our experiments reveal that while position bias remains noticeable in small-scale models and Non-English languages, increasing model scale and employing standard Chain-of-Thought prompting significantly mitigate this issue.
Moreover, we demonstrate that emerging \emph{thinking models} with extended reasoning capabilities are nearly immune to position bias, suggesting a costly but general approach toward resolving this vulnerability.
\end{abstract}

\section{Introduction}

Large language models (LLMs) have achieved remarkable success across a wide spectrum of natural language processing tasks, demonstrating capabilities that often rival or surpass human performance~\cite{brown2020language, ouyang2022training, comanici2025gemini, liu2025deepseek, yang2025qwen3, openai2025gpt5}. By scaling model parameters and training data, LLMs have evolved from simple statistical predictors into sophisticated systems capable of complex reasoning \cite{jaech2024openai, guo2025deepseek} and multilingual communication \cite{jiao2023chatgpt, wu-etal-2024-transagents}. 

Despite these advancements, LLMs continue to exhibit counter-intuitive vulnerabilities that undermine their reliability. One prominent issue is \emph{position bias}, where a model’s performance fluctuates significantly based on the sequential order of input information rather than its semantic content~\cite{mcilroy2024order, yu-etal-2025-mitigate, wang-etal-2025-position, brown2025order, wang2025eliminating, liu-etal-2024-lost}. Prior research has identified two primary manifestations of this phenomenon. In multiple-choice tasks, models often show systematic preferences for specific option labels (e.g., favoring option ``(B)'')~\cite{mcilroy2024order}. In knowledge-intensive and long-context tasks, models struggle to retrieve relevant documents placed in the middle of long input sequences, a phenomenon commonly referred to as \emph{Lost-in-the-Middle}~\cite{liu2024lost}. 

Position bias poses two major risks. First, it threatens the validity of LLM evaluation. Multiple-choice question answering (MCQA) benchmarks \cite{hendrycks2021measuring, zellers-etal-2019-hellaswag, suzgun2023challenging} remain a widely used standard for assessing general intelligence. If models achieve high scores by exploiting positional preference rather than genuine understanding, reported progress may be misleading. Second, severe position bias undermines real-world applicability, particularly in retrieval-augmented generation (RAG) and long-context systems \cite{lewis2020retrieval, ma2025blockattention, hsieh2024ruler}. If a model systematically favors information at specific positions, system accuracy and trustworthiness can be fundamentally compromised. 
Consequently, a thorough assessment of position bias is critical for conducting rigorous evaluations and drawing reliable conclusions within the LLM research.

However, a comprehensive picture of position bias in modern LLMs remains elusive, leaving it unclear how close current systems are to achieving permutation-invariant behavior. Existing literature exhibits several notable limitations. 
(1) \textit{Limited model scope}: many studies \cite{wang-etal-2025-position, mcilroy2024order, liu-etal-2024-lost} focus on earlier model families, such as LLaMA-2~\cite{touvron2023llama} and GPT-3.5 \cite{ouyang2022training}, and fail to capture the behavior of frontier models trained with more advanced architectures~\cite{liu2025deepseek, chung2024scaling} and higher-quality data \cite{wang2025octothinker}. 
(2) \textit{Neglect of new inference paradigms}: recent innovations, such as \emph{thinking models} capable of long-form reasoning (i.e., System~2-style inference)~\cite{guo2025deepseek, jaech2024openai}, represent a paradigm shift in model behavior, yet their impact on position bias remains largely unexamined. (3) \textit{English-centric evaluation}: most analyses are conducted exclusively in English, leaving typologically diverse languages (e.g., Arabic, Chinese, and Japanese) underexplored. 

In this work, we address these gaps through a systematic and large-scale investigation. We evaluate a diverse set of models spanning a wide range of scales, from Qwen3-1.7B to DeepSeek-V4 1.6T, across both multiple-choice and multi-document question answering tasks in five languages. Our experiments yield several key findings. First, we confirm that position bias is pronounced in small-scale models (i.e., fewer than $32$B parameters) and in non-English languages. However, increasing model size substantially alleviates this issue, with frontier models such as GPT-5.2 and DeepSeek-V4 exhibiting markedly greater robustness. Second, we show that prompting strategies play a critical role: even simple zero-shot Chain-of-Thought prompting consistently reduces position bias. Finally, we find that thinking models, such as DeepSeek-V3.2 with thinking mode enabled, are almost immune to position bias, suggesting that extended reasoning effectively decouples answer generation from positional bias.

Our contributions are summarized as follows:
\begin{itemize}
    \item We present a comprehensive evaluation of position bias across both multiple-choice and multi-document settings, spanning five languages (English, Chinese, Arabic, Spanish, and Japanese) and model scales from 1.7B parameters to frontier models such as  DeepSeek-V4 and GPT-5.2.
    \item We systematically analyze the factors influencing position bias, demonstrating that model scaling and standard Chain-of-Thought prompting significantly mitigate the issue.
    \item We show that the emerging class of thinking models, despite their computational cost, provides a robust solution to position bias, achieving near-uniform performance regardless of information ordering and pointing toward a general way for robust LLM design.
\end{itemize}

\section{Problem Formulation}

In this section, we formally define the task of quantifying position bias in Large Language Models (LLMs). 

\subsection{General Definition of Position Bias}
Given an input context $x$ sampled from a  dataset $\mathcal{D}$, the LLM $f_\theta$ generates an answer text $y$ for it.
We consider tasks where the input context $x$ is composed of a query $q$ and a set of $K$ information segments $\mathcal{S} = \{s_1, s_2, \dots, s_K\}$. These segments contain the necessary information to respond to $q$, e.g., retrieved documents \cite{lewis2020retrieval} or answer options \cite{hendrycks2021measuring}. Let $\pi$ be a permutation over the indices $\{1, \dots, K\}$. The input context under permutation $\pi$ is defined as:
$$
x_\pi = \phi(q, s_{\pi(1)}, s_{\pi(2)}, \dots, s_{\pi(K)}),
$$
where $\phi(\cdot)$ is a prompt function that linearizes these segments into a sequence. Ideally, an LLM should satisfy permutation invariance with respect to the semantic content of $\mathcal{S}$. That is, performance of $f_\phi$ should remain stable regardless of the order of the segments \cite{liu2024lost, wang2025eliminating}:
\begin{equation}
\sum_{x\sim\mathcal{D}}  M\big(f(x_\pi), y^*\big) \approx\sum_{x\sim\mathcal{D}} M\big(f(x_{\pi'}), y^*\big),
\end{equation}
where $M$ is the evaluation metric, e.g., exact-match (EM) score, $y^*$ is the golden truth for the input $x$, and $\pi$ and $\pi'$ are two different permutations for $\mathcal{S}$.

The \textbf{position bias} occurs when the model's performance fluctuates significantly based on the specific position $k$ assigned to the ground-truth segment $s^*$ within the linearized sequence.

\subsection{Multi-Choice Tasks}
Many fundamental evaluation for benchmarking the performance of LLMs are in the multi-choice form \cite{hendrycks2021measuring, zellers-etal-2019-hellaswag}. If most LLMs face a severe position bias issue in these tasks, a systematical reshaping for LLM evaluation may be necessary.

In the multi-choice (MC) setting, the LLM is presented with a question $q$ along with a set of candidate answer options $\mathcal{S}_{\textrm{MC}} = \{s_1, \dots, s_K\}$, typically labeled from ``(A)'' to ``(D)'', where position bias manifests as a preference for certain option orderings independent of answer content. Let $s^*$ denote the ground option. To analyze position bias, we shift the position $k$ of $s^*$ across all positions in $\{1, \dots, K\}$, yielding a permutation set $\mathcal{P}_{\textrm{MC}} = \{\pi^1, \dots, \pi^K\}$, where $\pi^k$ is a permutation of $\{1, \dots, K\}$ with the golden answer at the $k$-th position, i.e., $s_k = s^*$.

\subsection{Multi-Doc Tasks}
Correctly using the information of documents in the context is vital for many LLM applications, e.g., retrieval-augmented generation \cite{lewis2020retrieval} and long-context modeling \cite{hsieh2024ruler}. Investigating the position bias in this kind of tasks is also important.

Different from the the multi-choice tasks, where the options in $\mathcal{S}_{\mathrm{MC}}$ are generally short and close to the question, the multi-doc tasks integrate long and independent documents into the context. In this setting, the LLM is presented with a set of long documents $\mathcal{S}_{\textrm{MD}} = \{s_1, \dots, s_K\}$ before the question $q$, where only one golden document $s^* \in \mathcal{S}$ contain the answer for $q$. We shift the position $k$ of $s^*$ across sampled positions similarly, resulting a permutation set ${P}_{\textrm{MD}}$. 

\subsection{Evaluation Metrics}
The main evaluation metric $M$ we used to measure the performance of model $f_\phi$ is the exact-match (EM) \cite{liu2024lost, wang2025eliminating, hsieh2024ruler} under a specific permutation $\pi$.
\begin{equation}
e_{\pi,\theta} = \frac{1}{|\mathcal{D}|}  \sum_{x\sim\mathcal{D}}  M\big(f(x_\pi), y^*\big).
\end{equation}
We also report the largest performance difference of $f_\phi$ is at different positions:
\begin{equation}
    \Delta_{\theta} = \max_{\pi\in\mathcal{P}} e_{\pi,\theta} - \min_{\pi\in\mathcal{P}} e_{\pi,\theta},\label{eq:delta}
\end{equation}
where $\mathcal{P}$ is the permutation set.

\section{Multi-Choice Question Answering\label{sec:mcqa}}

\begin{figure*}[htbp]
    \centering
    \begin{subfigure}[b]{0.48\textwidth}
        \centering
        \includegraphics[width=\textwidth]{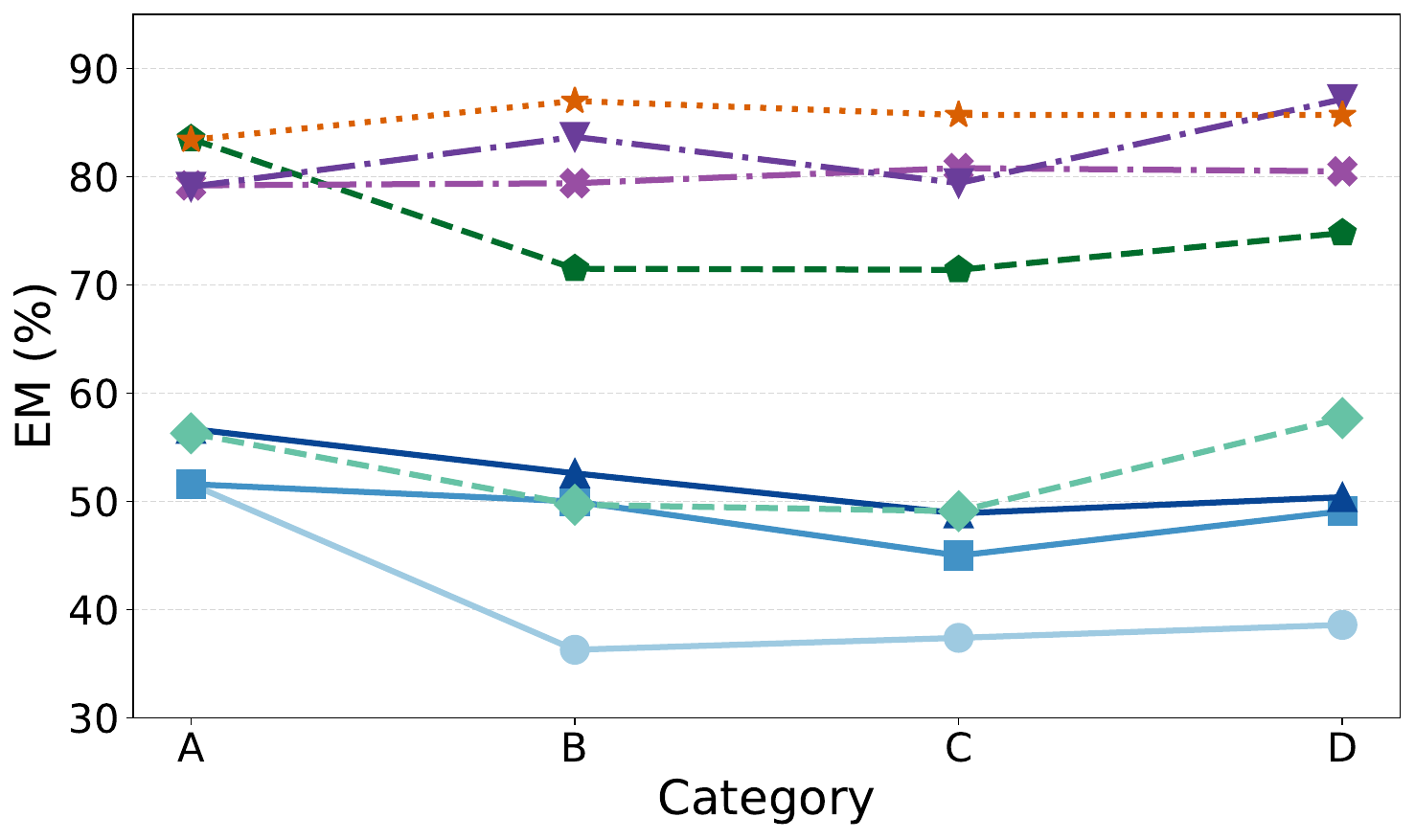} 
        \caption{0-shot DA}
        \label{fig:en_0shot_da}
    \end{subfigure}
    \hfill 
    \begin{subfigure}[b]{0.48\textwidth}
        \centering
        \includegraphics[width=\textwidth]{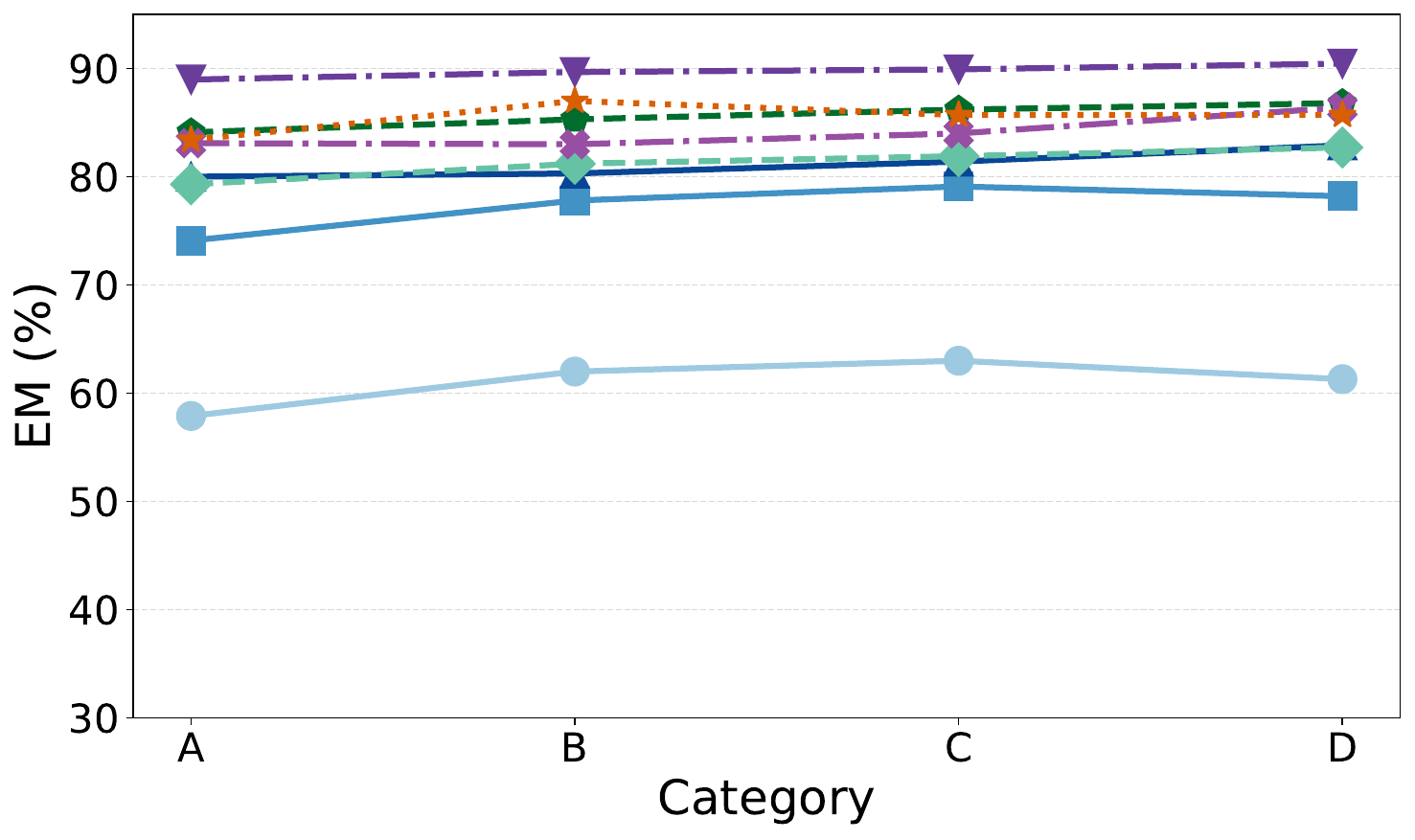}
        \caption{0-shot CoT}
        \label{fig:en_0shot_cot}
    \end{subfigure}

    \begin{subfigure}[b]{0.48\textwidth}
        \centering
        \includegraphics[width=\textwidth]{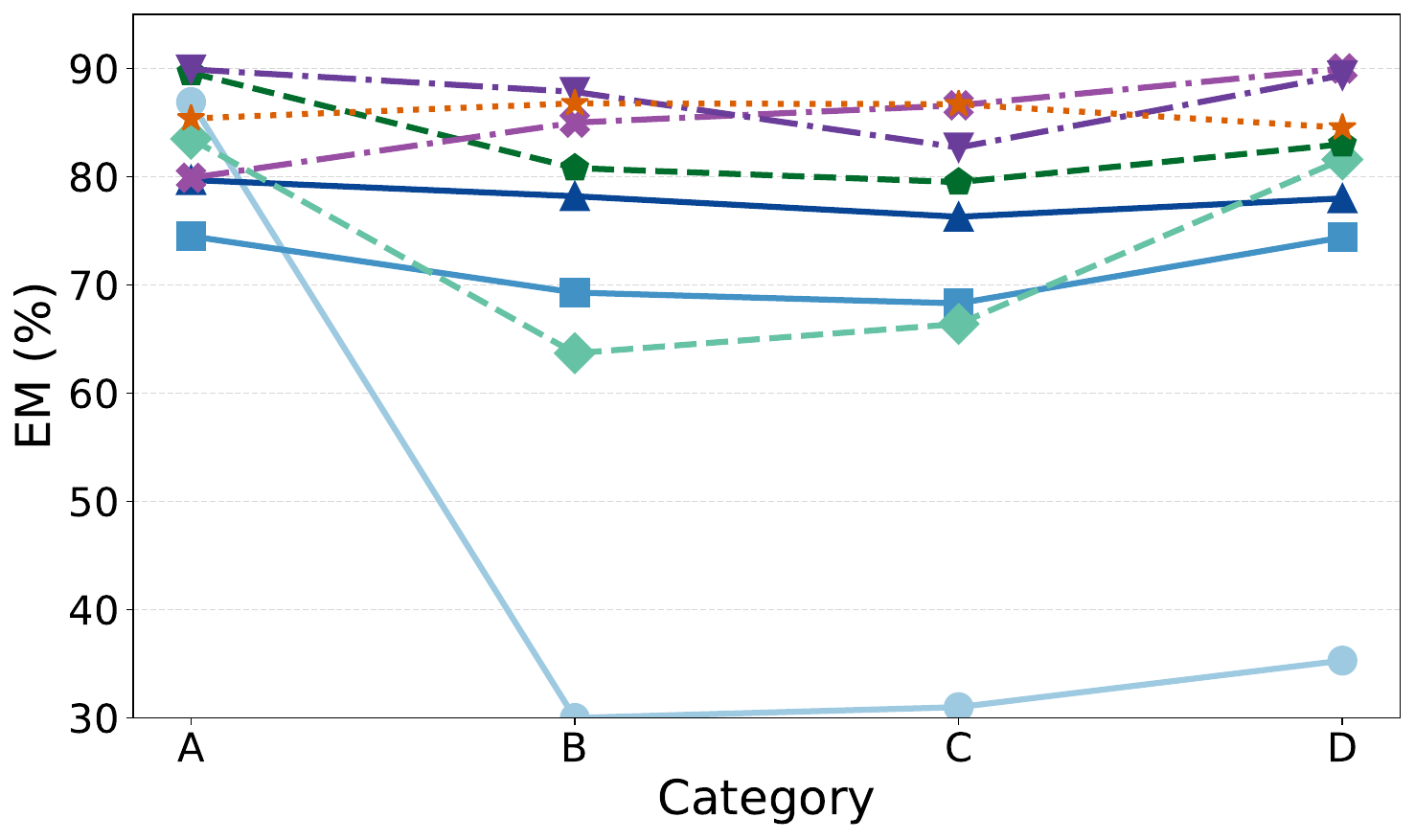}
        \caption{5-shot DA}
        \label{fig:en_5shot_da}
    \end{subfigure}
    \hfill
    \begin{subfigure}[b]{0.48\textwidth}
        \centering
        \includegraphics[width=\textwidth]{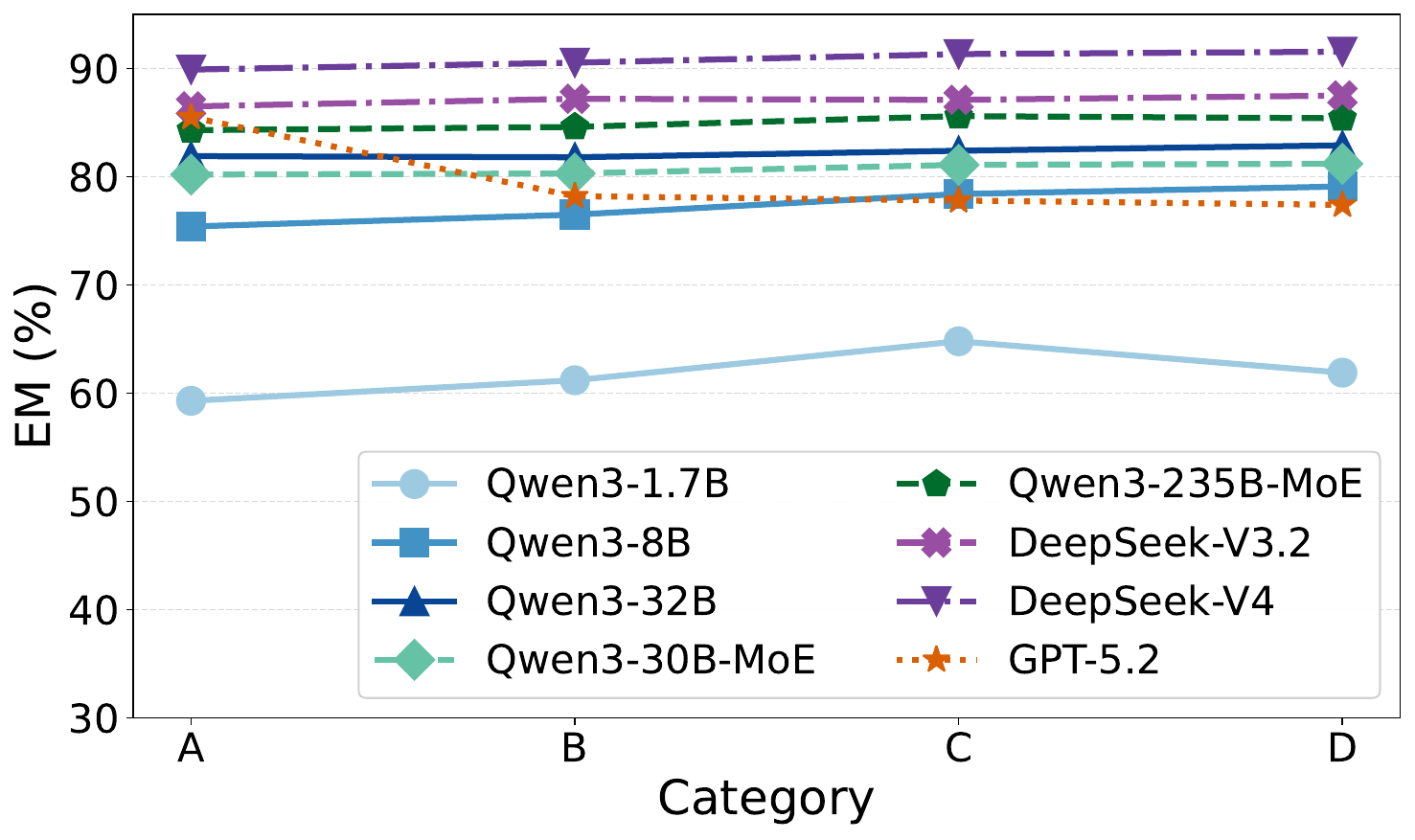}
        \caption{5-shot CoT}
        \label{fig:en_5shot_cot}
    \end{subfigure}
    
    \caption{Position bias on English Multi-Choice tasks.  
    Results are reported on  MMLU benchmark \cite{hendrycks2021measuring} using non-thinking models. The label of golden answer is shifted from ``(A)'' to ``(D)'' for all data, and the options are positioned in alphabetical order of the labels, i.e., ``(A)'' appearing first and ``(D)'' last.}
    \label{fig:mmlu}
\end{figure*}

\subsection{Setup}
The MMLU dataset\footnote{\url{https://huggingface.co/datasets/cais/mmlu}} \cite{hendrycks2021measuring}, which comprises 57 subtasks, is widely used to benchmark the general intelligence of large language models (LLMs). Each question includes four answer options labeled ``(A)'' through ``(D)'', only one of which is correct. In the original benchmark, the position of the correct answer is randomly shuffled to ensure an even distribution across all option positions.

In contrast, in our experiments, we evaluate the position bias of LLMs by forcing the correct answer to appear at a fixed position (one of ``(A)'' to ``(D)'') for all questions. The re-labeled option list is positioned in alphabetical order of the labels, with option ``(A)'' appearing first and ``(D)'' last. We then visualize and compare the model’s performance across runs with different fixed answer positions. This procedure yields $14{,}052 \times 4 = 56{,}168$ test examples per run on MMLU benchmark, covering all 57 subtasks. 

In addition to evaluation on the MMLU dataset in English, we are also curious about the position bias in non-English settings. Therefore, we leverage the MMMLU\footnote{\url{https://huggingface.co/datasets/openai/MMMLU}}, which is parallel to MMLU, to evaluate on four additional languages: Spanish (alphabetic and typologically closer to English), Arabic (right-to-left script), and Chinese and Japanese (representing logographic and mixed-script systems). We sample 25 data points in each sub-task, resulting $25\times57\times4=5,700$ test examples\footnote{Preliminary study on English data show that evaluation on the down-sampled dataset yields consistent observations as the those on full dataset.} for each Non-English language. 

In our experiments, we consider four commonly used evaluation settings:
\begin{description}
    \item[0-shot DA] The model is prompted to directly answer (DA) by selecting an option label.
    \item[5-shot DA] We include the 5-shot DA examples provided by \citet{chung2024scaling} in the prompt.
    \item[0-shot CoT] We add the text ``Let’s think step by step'' into the prompt to elicit Chain-of-Thought (CoT) reasoning \cite{wei2022chain} before producing the final answer.
    \item[5-shot CoT] We include 5-shot CoT examples in the prompt to enhance the reasoning process, following \citet{chung2024scaling}.
\end{description}

To systematically assess the severity of position bias, we conduct experiments on models spanning a wide range of scales: Qwen3-1.7B, Qwen3-8B, Qwen3-32B, Qwen3-30B-MoE, Qwen3-235B-MoE \cite{yang2025qwen3}, DeepSeek-V3.2 with 685B parameters \cite{liu2025deepseek}, DeepSeek-V4 1.6T and GPT-5.2\footnote{\url{https://platform.openai.com/docs/models/gpt-5.2}}. We use the official APIs provided by Aliyun\footnote{\url{https://modelstudio.console.alibabacloud.com/}}, DeepSeek\footnote{\url{https://api-docs.deepseek.com}}, and OpenAI\footnote{\url{https://platform.openai.com/docs/models/gpt-5.2}} to access Qwen3, DeepSeek V3.2 and V4, and GPT-5.2, respectively.
For all models, we set the temperature to 0 during generation and evaluate performance using the exact-match score between the predicted and golden-truth option labels.  

\begin{table}[]
  \resizebox{1.0\columnwidth}{!}{
  \begin{tabular}{lccccc}\toprule
  \multicolumn{1}{l}{} & 0shot & 5shot & 0shot CoT & 5shot CoT & AVG. \\
  \midrule
  Q3-1.7B        & 15.3 & 56.9 & 5.1 & 5.5 & 20.7 \\
  Q3-8B          & 6.6  & 6.2  & 5.0 & 3.7 & 5.4 \\\hdashline
  Q3-30B         & 8.6  & 19.8 & 3.4 & \textbf{1.0} & 8.2 \\
  Q3-32B         & 7.8  & 3.4  & 2.9 & 1.1 & \textbf{3.8} \\\hdashline
  Q3-235B        & 12.1 & 10.1 & 2.7 & 1.3 & 6.6 \\
  DS-V3.2        & \textbf{1.6} & 10.1 & 3.4 & \textbf{1.0} & 4.0 \\
  DS-V4      & 8.1  & 7.2  & \textbf{1.5} & 1.7 & 4.6 \\
  GPT-5.2        & 3.6  & \textbf{2.2} & 3.6 & 8.1 & 4.4 \\\bottomrule
  \end{tabular}
  }\caption{The performance difference $\Delta_\theta$ of LLMs on English
  Multi-Choice tasks.
  Results are reported on MMLU using non-thinking models.
  }\label{tab:mmlu_delta}
  \end{table}

\subsection{Main Results}
As shown in Figure \ref{fig:mmlu}, we compare model performance when the golden truth is forced to appear in only one of the four options. We also report the largest performance difference (Equation \ref{eq:delta}) for each model in Table \ref{tab:mmlu_delta}.

\paragraph{Impact of Model Scale}
In many previous works, similar evaluations have mainly been conducted on a limited range of models \cite{wang2025eliminating, brown2025order, reif-schwartz-2024-beyond}. However, it is important to understand how severe the position bias issue is across models of different scales.

We observe that the small-scale Qwen3-1.7B fluctuates significantly when the position of the golden answer changes. For example, the EM score drops by 56.9 points when the position changes from ``(A)'' to ``(B)'' in Figure \ref{fig:en_5shot_da}. This non-uniform performance is noticeably alleviated in mid-scale models, such as Qwen3-32B. Moreover, the two largest models, i.e., Deepseek-V4-Pro and GPT5.2, achieve much more stable performance across all four positions. These observations demonstrate that model capacity is strongly related to the severity of position bias; specifically, scaling up model size can significantly mitigate this issue.

\begin{figure*}[htbp]
    \centering
    \begin{subfigure}[b]{0.48\textwidth}
        \centering
        \includegraphics[width=\textwidth]{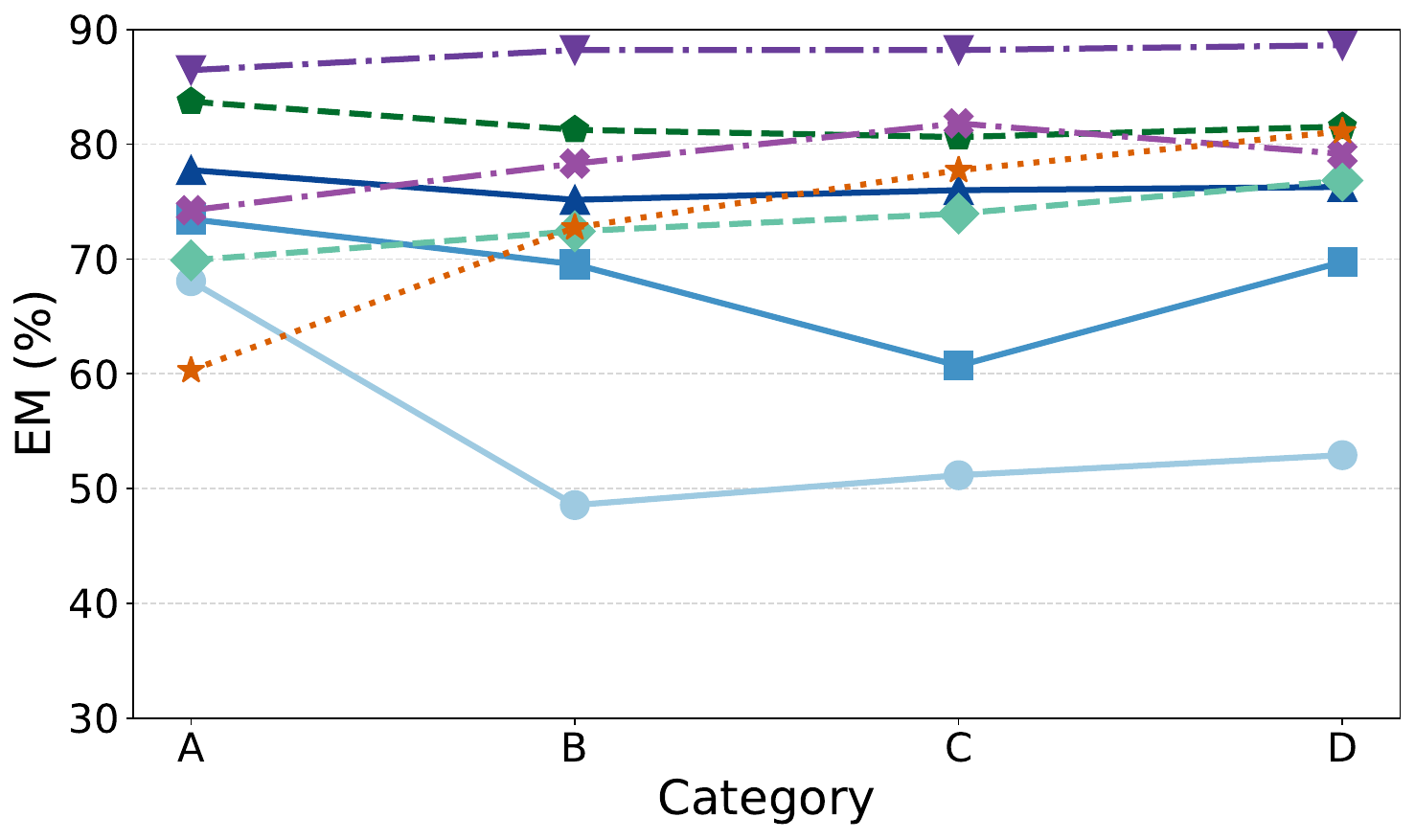} 
        \caption{Chinese}
        \label{fig:zh_0shot_cot}
    \end{subfigure}
    \hfill
    \begin{subfigure}[b]{0.48\textwidth}
        \centering
        \includegraphics[width=\textwidth]{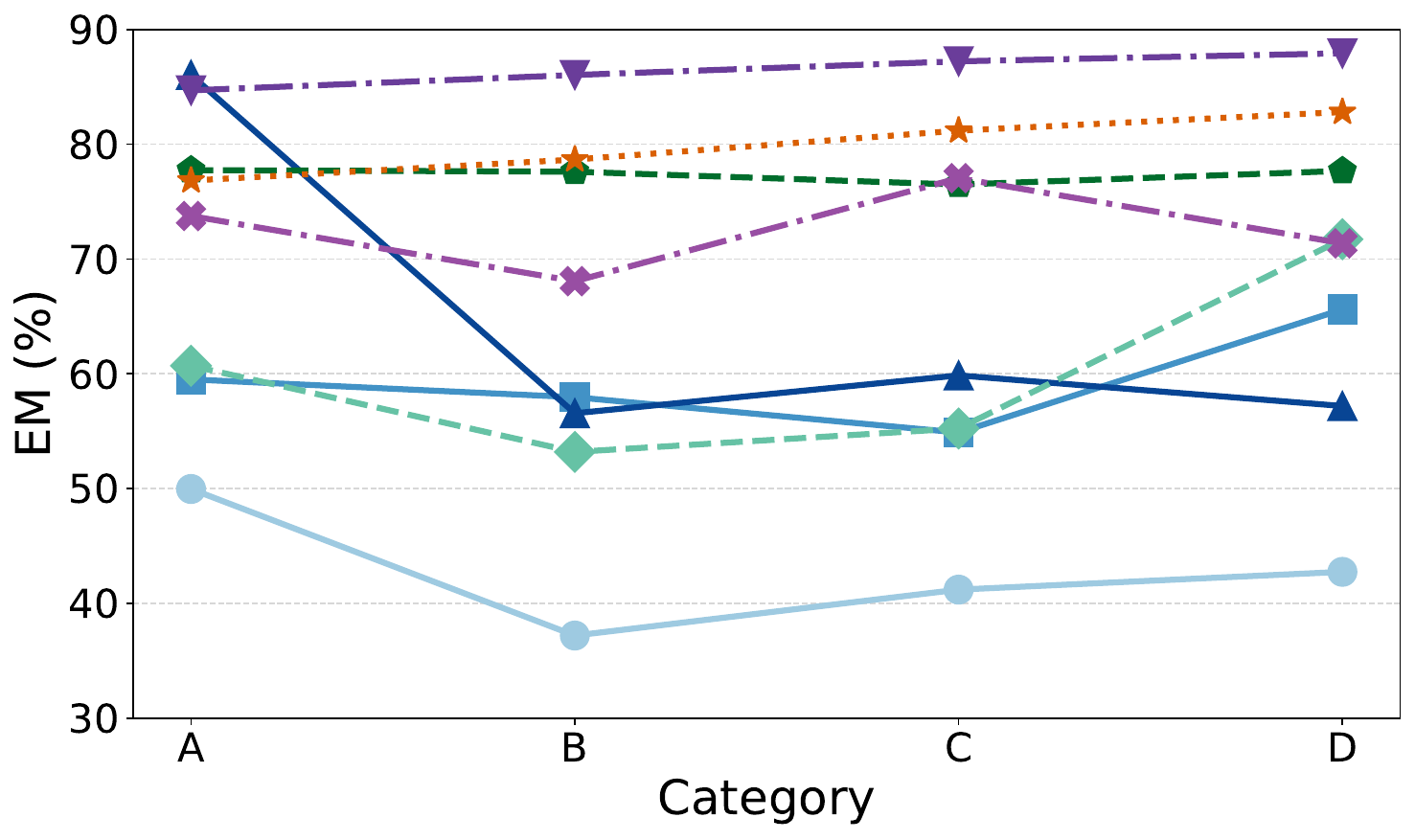}
        \caption{Arabic}
        \label{fig:ar_0shot_cot}
    \end{subfigure}
    
    \begin{subfigure}[b]{0.48\textwidth}
        \centering
        \includegraphics[width=\textwidth]{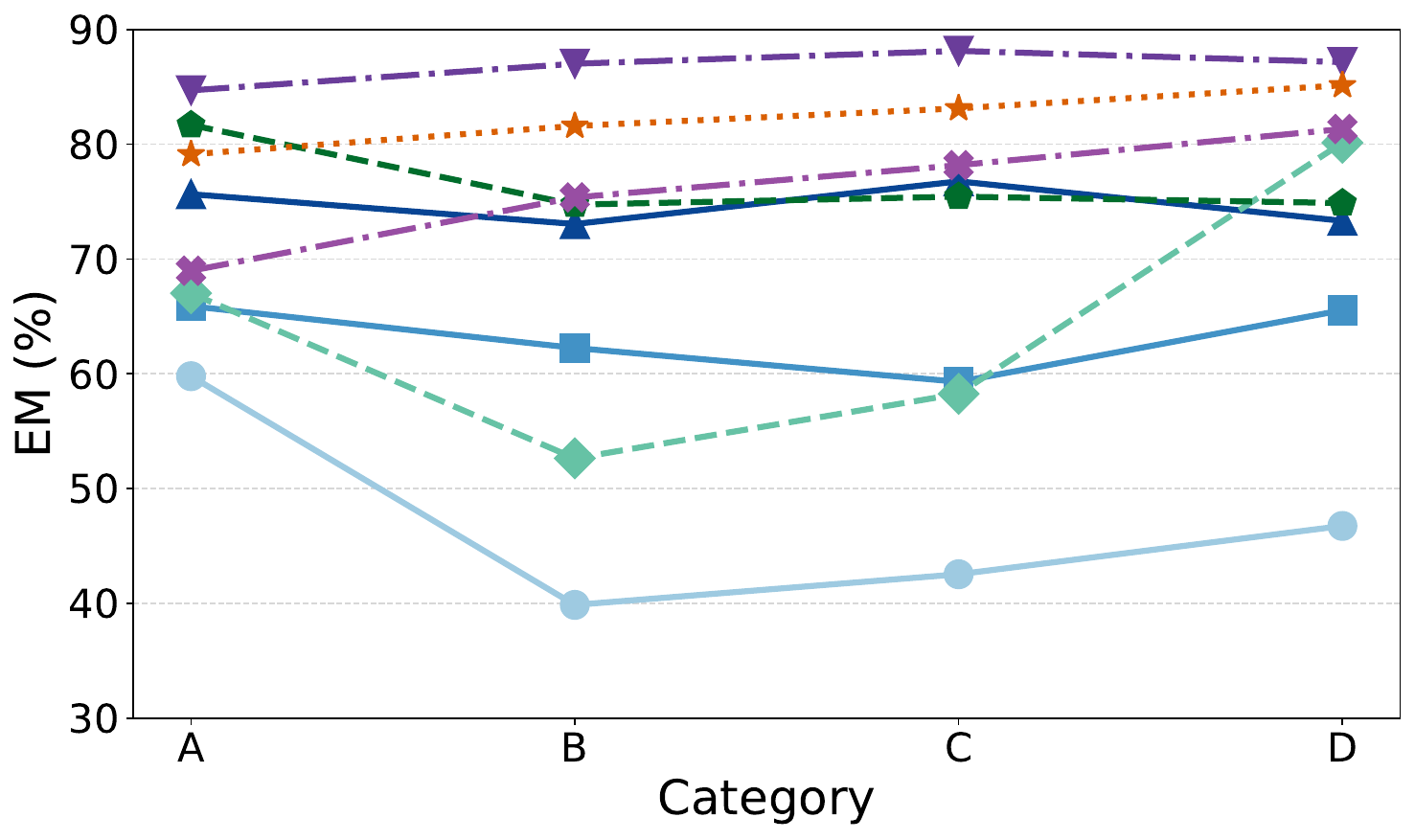}
        \caption{Japanese}
        \label{fig:ja_0shot_cot}
    \end{subfigure}
    \hfill
    \begin{subfigure}[b]{0.48\textwidth}
        \centering
        \includegraphics[width=\textwidth]{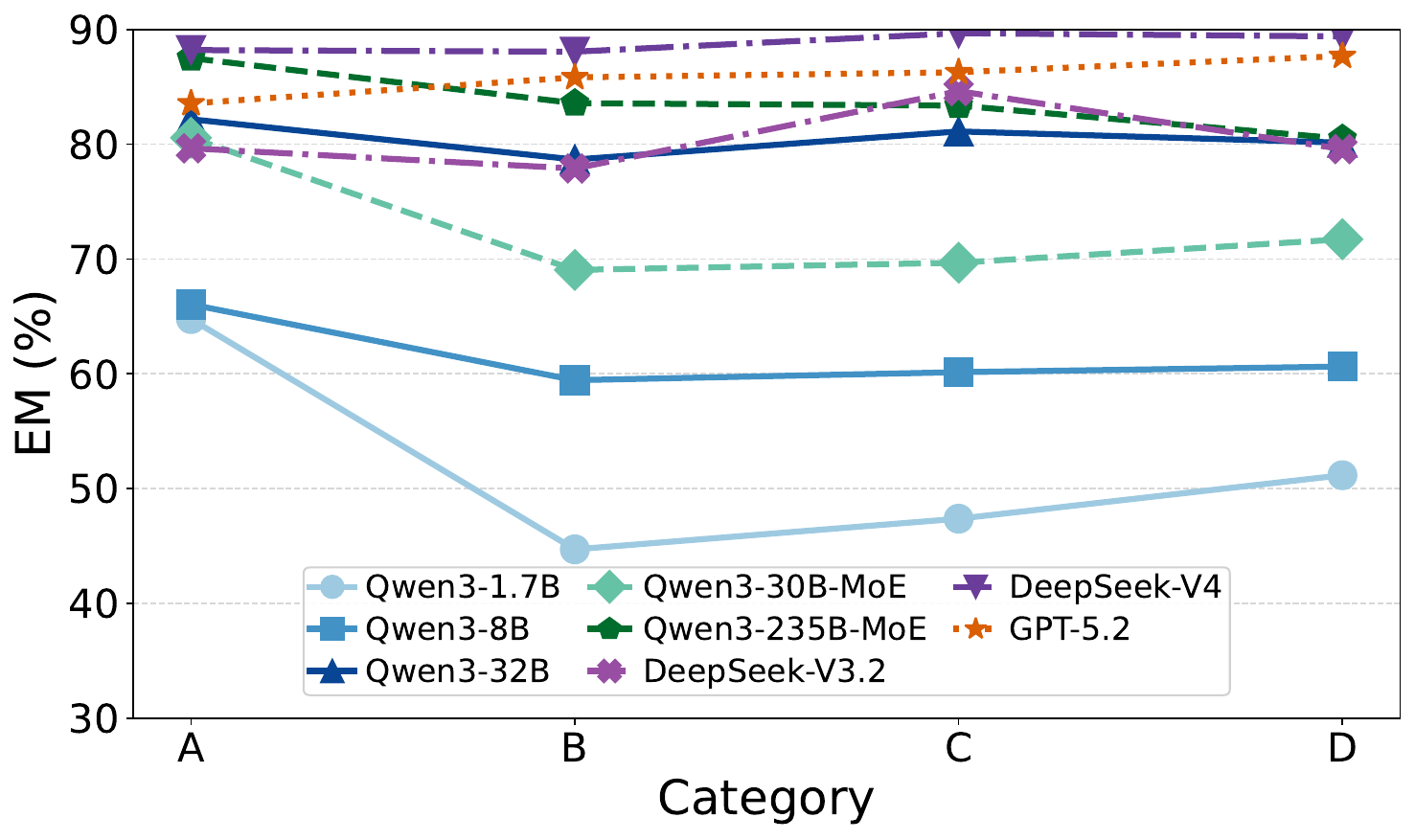}
        \caption{Spanish}
        \label{fig:es_0shot_cot}
    \end{subfigure}
    
    \caption{Position bias on Multilingual Multi-Choice tasks. Results are reported on MMMLU benchmark using non-thinking models and 0-shot CoT template. The label of golden answer is shifted from ``(A)'' to ``(D)'' for all data, and the options are positioned in alphabetical order of the labels, i.e., ``(A)'' appearing first and ``(D)'' last.}
    \label{fig:mmmlu}
\end{figure*}

\paragraph{Impact of Few-Shot Examples} 
Few-shot examples are widely used to guide models toward generating higher-quality answers. However, the few-shot examples themselves may introduce additional position bias; for instance, the option labels of golden answers may not be evenly distributed~\cite{chung2024scaling}. 

While most models improve overall in the 5-shot DA setting (Figures \ref{fig:en_0shot_da}, \ref{fig:en_5shot_da}), Qwen3-1.7B, Qwen3-30B-MoE, and even Deepseek-V3.2 exhibit more severe position bias when using the 5-shot DA examples provided by FLAN \cite{chung2024scaling}. This suggests that position bias may not originate solely from the model, but can also be influenced by the information provided in the context. But, the influence of few-shot examples on frontier models with trillion parameters, such as Deepseek V4 and GPT-5.2, is marginal.

\paragraph{Impact of CoT} 
Chain-of-Thought (CoT) prompting is widely used in practice and may alleviate position bias by encouraging the model to reason more thoroughly before answering. 

As shown in Figure \ref{fig:en_0shot_cot}, simply adding the phrase ``Let's think step by step'' to the prompt significantly reduces position bias for almost all models, resulting in a more uniform performance distribution across different option labels. In Figure \ref{fig:en_5shot_cot}, we observe consistently low-level position bias when 5-shot examples with CoT are provided. However, the 5-shot CoT examples do not improve overall performance, indicating that the modern LLMs have learned high-quality reasoning abilities.


\begin{table}[]
  \centering
  \resizebox{1.0\columnwidth}{!}{
  \begin{tabular}{lcccccc}\toprule
  \multicolumn{1}{l}{} & EN &  ZH & AR & JA & ES & AVG. \\\midrule
  Q3-1.7B        & 5.1   & 19.5   & 12.8  & 19.9    & 20.1    & 18.1 \\
  Q3-8B          & 5.0   & 12.8   & 10.7  & 6.6     & 6.6     & 9.2 \\
  \hdashline
  Q3-30B         & 3.4   & 7.0    & 18.5  & 27.5    & 11.5    & 16.1 \\
  Q3-32B         & 2.9   & 2.6    & 29.5  & 3.7     & 3.5     & 9.8 \\
  \hdashline
  Q3-235B        & 2.7   & 3.1    & \textbf{1.3} & 6.9 & 7.0 & 4.6 \\
  DS-V3.2        & 3.4   & 7.6    & 9.0   & 12.4    & 6.7     & 8.9 \\
  DS-V4      & \textbf{1.5} & \textbf{2.2} & 3.2 & \textbf{3.4} &
  \textbf{1.6} & \textbf{2.6} \\
  GPT-5.2        & 3.6   & 20.8   & 6.0   & 6.0     & 4.1     & 9.2 \\
  \bottomrule
  \end{tabular}
  }\caption{The performance difference $\Delta_\theta$ of LLMs on Multilingual
  Multi-Choice tasks. Results are reported on MMMLU benchmark using non-
  thinking models and 0-shot CoT prompt.}
  \label{tab:mmmlu_delta}
  \end{table}
  
\subsection{Results on Multilingual Data}
Prior work on position bias has focused primarily on English \cite{wang2025eliminating, wang-etal-2025-position, tripathi2025pairwise, tian-etal-2025-distance}, leaving its severity in other languages underexplored. We therefore examine four languages: Spanish (similar to English), Arabic (right-to-left order), and Chinese and Japanese (logographic and mixed-script systems).

As shown in Figure \ref{fig:mmmlu} and Table \ref{tab:mmmlu_delta}, even with the 0-shot CoT method, many LLMs exhibit severe position bias on non-English data. Specifically, the Qwen3-30B-MoE model, which achieves almost uniform performance in English (Figure \ref{fig:en_0shot_cot}), shows noticeable performance fluctuations in Arabic, Japanese, and Spanish. Even the Deepseek-V3.2 and GPT-5.2 models demonstrate more pronounced position bias in non-English languages.

\subsection{Results on the Thinking Model}
The \emph{thinking model} \cite{guo2025deepseek, jaech2024openai}, which generates long-form reasoning trajectory before answering, has achieved considerable success. Since 0-shot CoT has shown positive effects on English data, we use the thinking model to evaluate whether enhanced CoT can further alleviate the position bias issue. 

\begin{figure}[!tbp]
    \centering
    \begin{subfigure}[b]{0.48\textwidth}
        \centering
        \includegraphics[width=\textwidth]{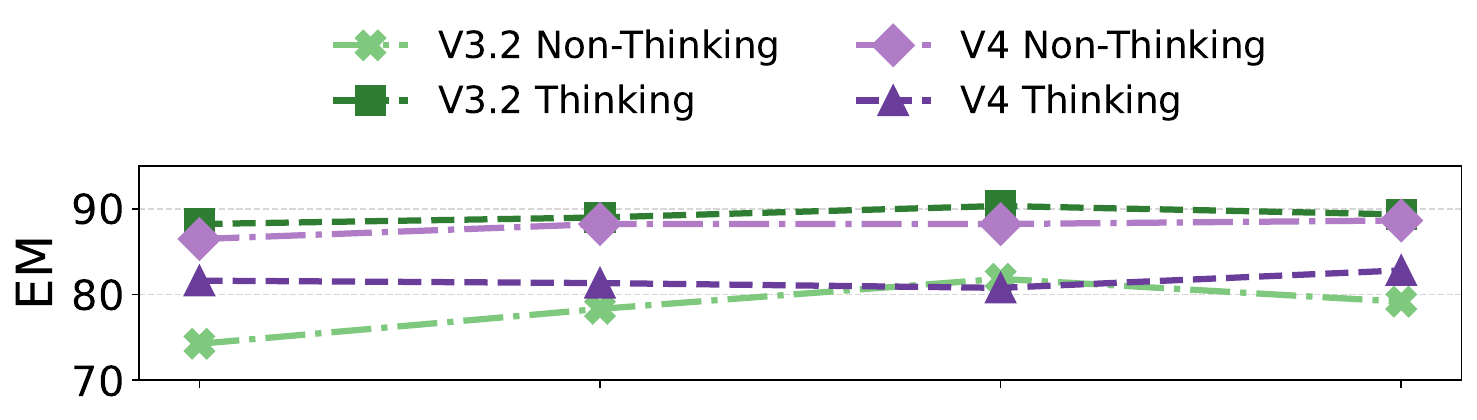} 
        \caption{Chinese}
        \label{fig:ds_zh_0shot_cot}
    \end{subfigure}
    \begin{subfigure}[b]{0.48\textwidth}
        \centering
        \includegraphics[width=\textwidth]{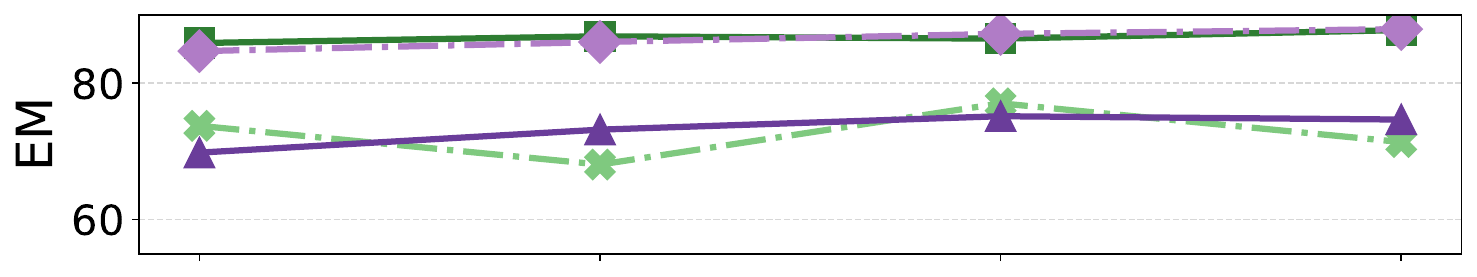}
        \caption{Arabic}
        \label{fig:ds_ar_0shot_cot}
    \end{subfigure}
    \begin{subfigure}[b]{0.48\textwidth}
        \centering
        \includegraphics[width=\textwidth]{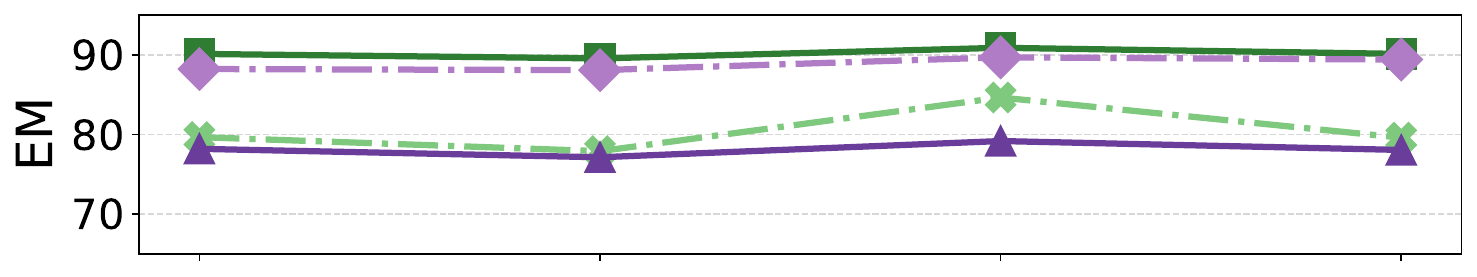}
        \caption{Spanish}
        \label{fig:ds_es_0shot_cot}
    \end{subfigure}
    \begin{subfigure}[b]{0.48\textwidth}
        \centering
        \includegraphics[width=\textwidth]{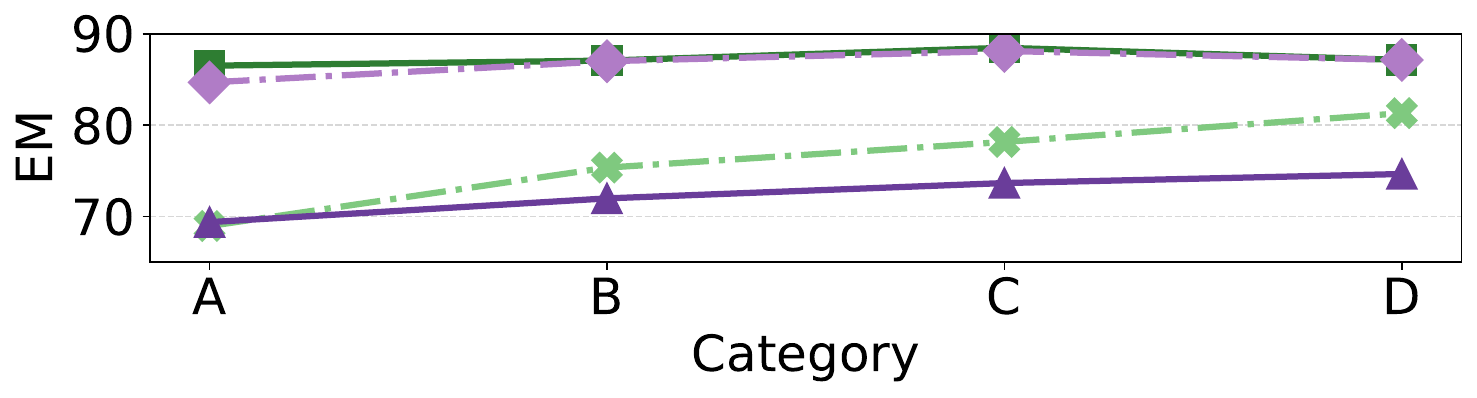}
        \caption{Japanese}
        \label{fig:ds_ja_0shot_cot}
    \end{subfigure}
    
    \caption{Position bias on Multilingual Multi-Choice tasks. We use the Deepseek-V3.2, Deepseek-V4-Pro and the 0-shot CoT tempalte for evaluation.}
    \label{fig:mmmlu_ds_thinking}
\end{figure}

As shown in Figure \ref{fig:mmmlu_ds_thinking}, Deepseek-V3.2-Thinking exhibits less position bias than its standard counterpart. Since Deepseek-V4 shows nearly uniform performance even without reasoning, there has  little room to further reduce the position bias with thinking mode.  Interestingly, DeepSeek V4 achieves a lower EM score in thinking mode compared to non-thinking mode. We hypothesize that this is driven by the "over-thinking" phenomenon on simple questions \cite{aggarwal2025optimalthinkingbench}.

\section{Multi-Document Tasks}

The \emph{Lost-in-the-Middle} (LITM) effect \cite{liu2024lost} describes the counter-intuitive performance degradation observed when relevant document is placed in the middle of a prompt. In this section, we comprehensively evaluate on modern LLMs to determine whether this phenomenon remains a significant challenge. 

\subsection{Setup}

\begin{figure*}[htbp]
    \centering
    \begin{subfigure}[b]{0.48\textwidth}
        \centering
        \includegraphics[width=\textwidth]{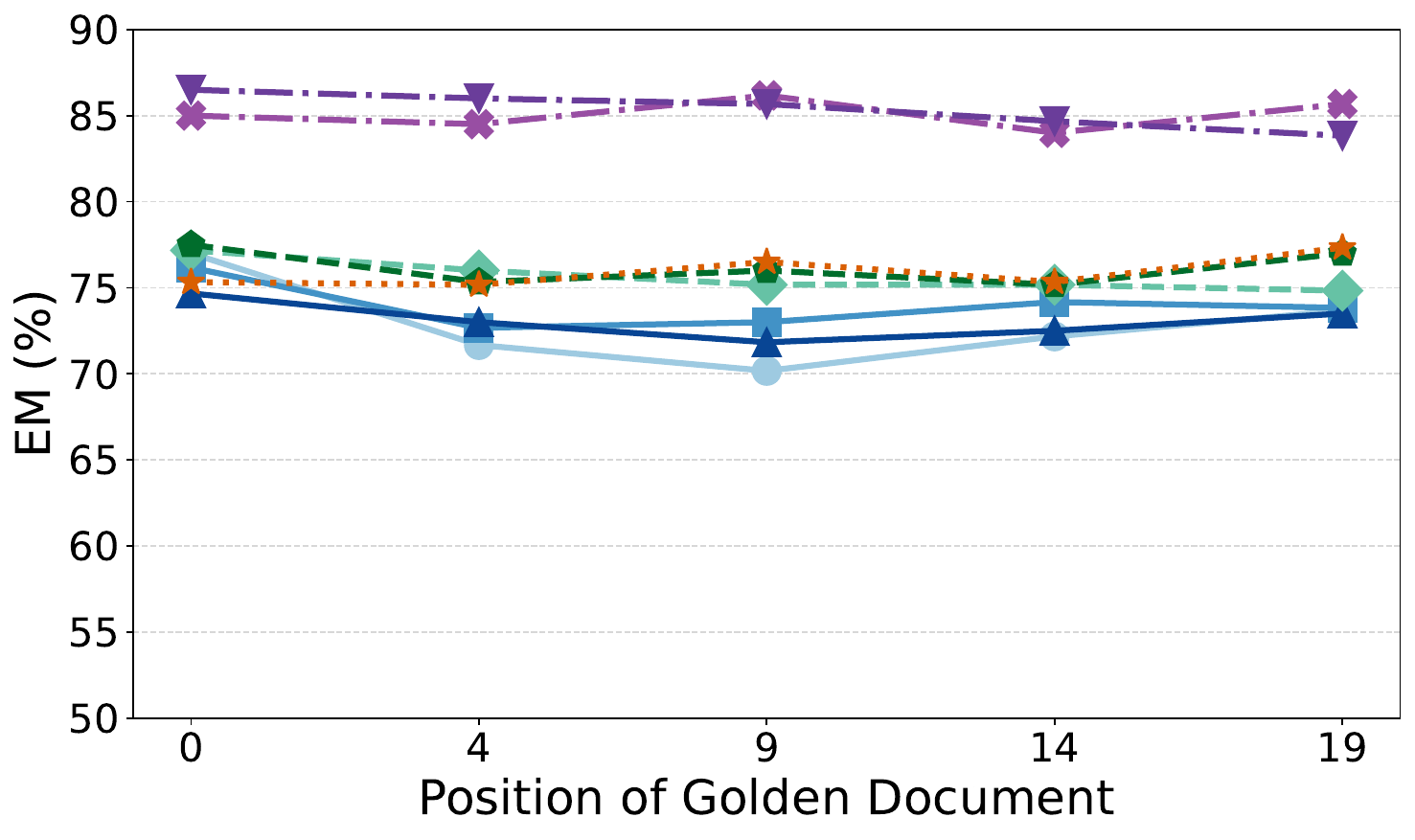} 
        \caption{English}
        \label{fig:qa_en}
    \end{subfigure}
    \begin{subfigure}[b]{0.48\textwidth}
        \centering
        \includegraphics[width=\textwidth]{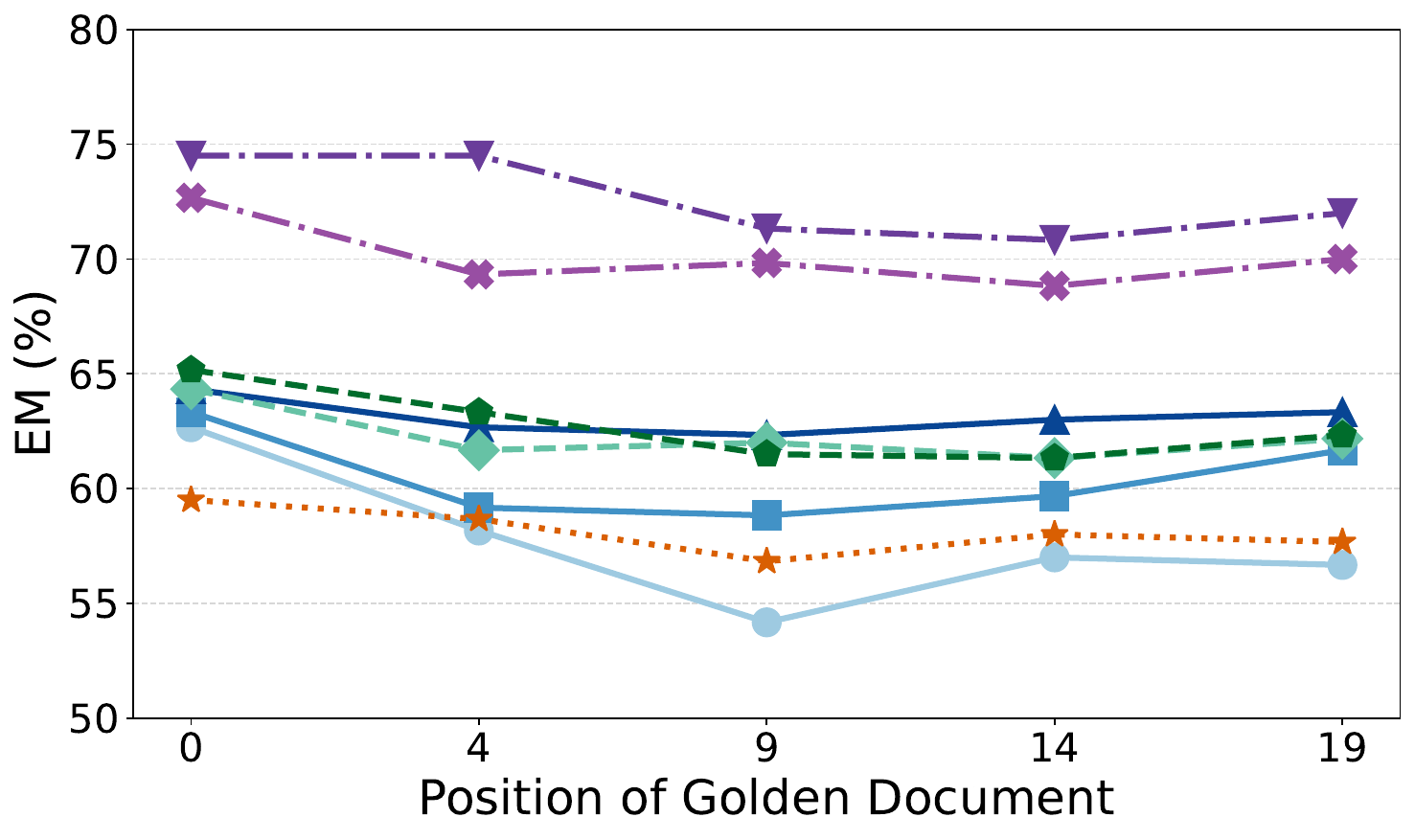}
        \caption{Chinese}
        \label{fig:qa_zh}
    \end{subfigure}
    \begin{subfigure}[b]{0.48\textwidth}
        \centering
        \includegraphics[width=\textwidth]{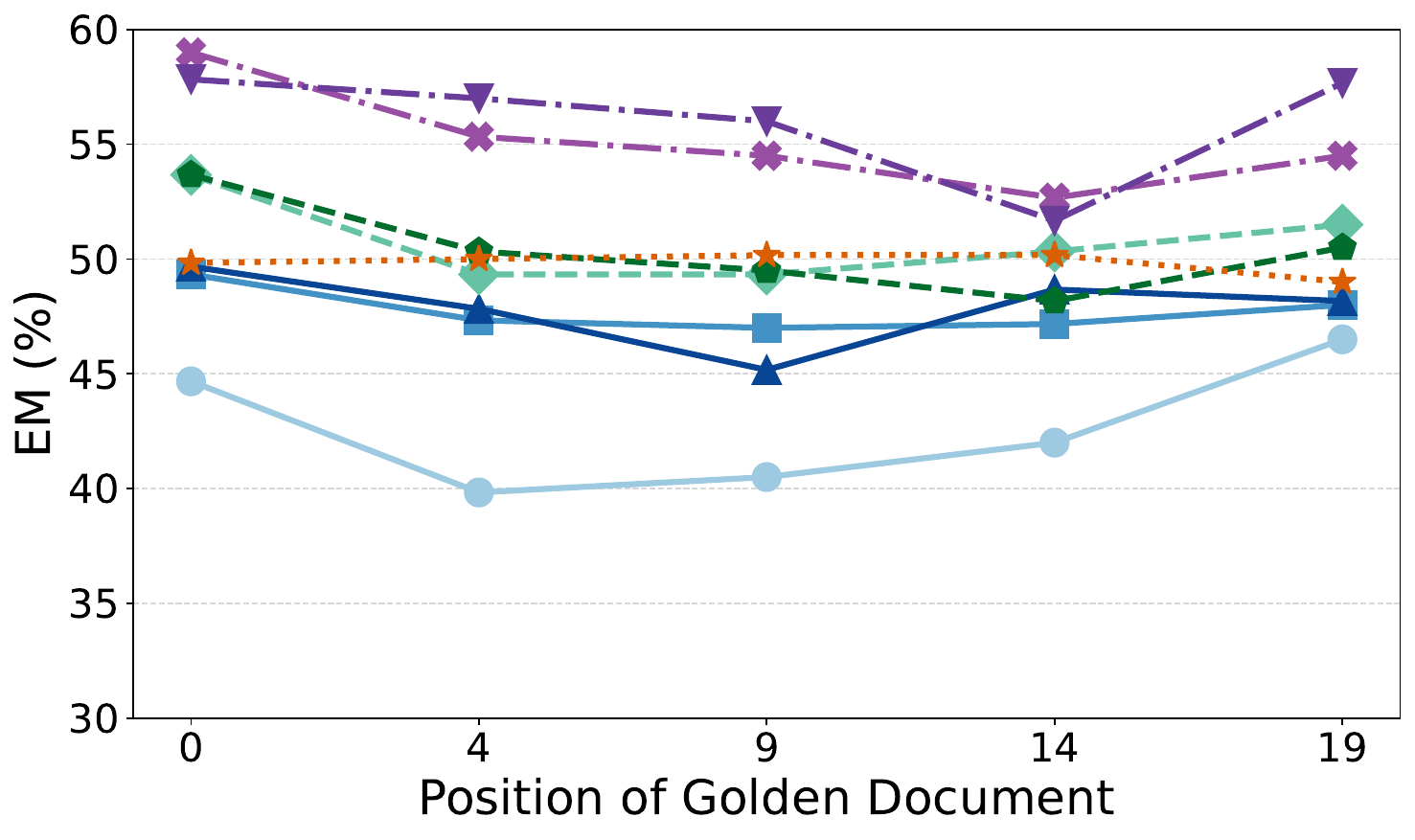}
        \caption{Arabic}
        \label{fig:qa_ar}
    \end{subfigure}
    \begin{subfigure}[b]{0.48\textwidth}
        \centering
        \includegraphics[width=\textwidth]{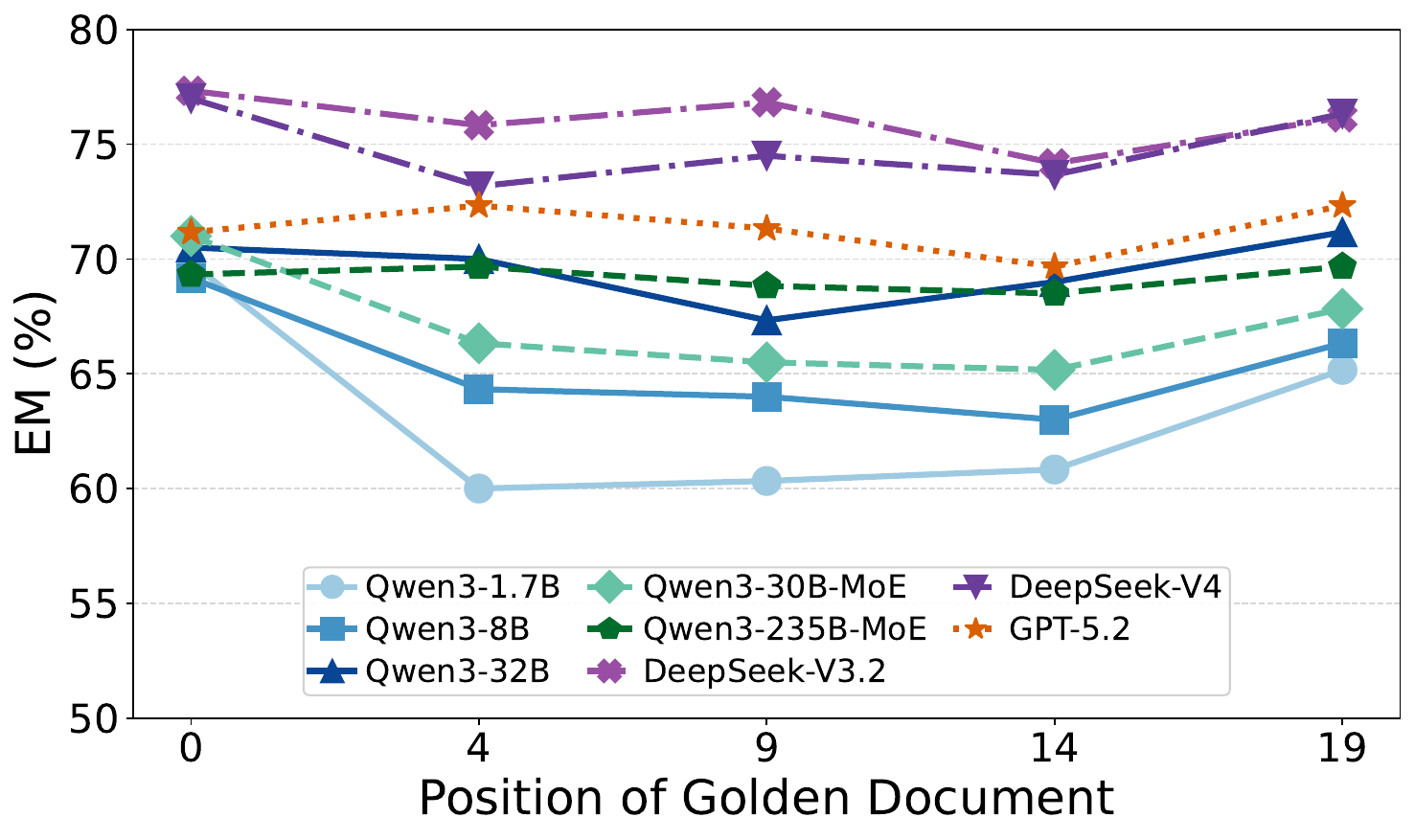}
        \caption{Spanish}
        \label{fig:qa_es}
    \end{subfigure}
    
    \caption{Position bias on Multi-Document QA. }
    \label{fig:qa}
\end{figure*}

 We use the MLQA dataset\footnote{\url{https://huggingface.co/datasets/facebook/mlqa}} \cite{lewis2020mlqa} to analyze whether the effect persists cross-linguistically.  Following the setup of the Lost in the Middle dataset\footnote{\url{https://github.com/nelson-liu/lost-in-the-middle}} provided by \citet{liu2024lost}, we construct N=20 documents per question with golden documents placed at positions 0, 4, 9, 14, and 19, sampling 500 instances per language (2,500 examples per run). Details of the dataset construction are provided in Appendix~\ref{app:data_construction}.

Following the experiments in Section \ref{sec:mcqa}, we evaluate models across a variety of scales: Qwen3-1.7B, Qwen3-8B, Qwen3-32B, Qwen3-30B-MoE, Qwen3-235B-MoE \cite{yang2025qwen3}, DeepSeek-V3.2, \cite{liu2025deepseek}, DeepSeek-V4 1.6T, and GPT-5.2. Following \citet{liu2024lost}, we use Exact Match (EM) as our primary evaluation metric.

\subsection{Main Results}

\paragraph{Impact of Model Scale} 
As shown in Table \ref{tab:qa_delta}, we calculate the largest performance difference ($\Delta_\theta$), as defined in Eq. \ref{eq:delta}, across all positions for each model. The overall trend indicates that $\Delta_\theta$ values decrease as the number of model parameters increases, although a few outliers exist. Many small-scale models still exhibit strong position bias; for instance, the performance differences for Qwen3-1.7B and Qwen3-8B are 9.8 and 6.2 points in Spanish, respectively. In contrast, the performance differences for large-scale models, such as Qwen3-235B-MoE, DeepSeek V3.2 and V4, and GPT-5.2, are maintained within 5 points across most languages other than Arabic.

\begin{table}[!ht] 
\centering
\begin{tabular}{lccccc}
\toprule
\multicolumn{1}{l}{} & EN & ZH & AR & ES & AVG. \\
\midrule
Q3-1.7B           & 6.8   & 8.5    & 6.7   & 9.8     & 8.0 \\
Q3-8B             & 3.5   & 4.5    & 2.3   & 6.2     & 4.1 \\
\hdashline
Q3-30B            & 2.3    & 3.0    & 4.3   & 5.8     & 3.9 \\
Q3-32B            & 2.8    & \textbf{2.0}    & 4.5   & 3.8     & 3.3 \\
\hdashline
Q3-235B           & 2.3   & 3.8    & 5.5   &\textbf{ 1.2}     & 3.2 \\
DS-V3.2           & \textbf{2.1}    & 3.8    & 6.3   & 3.2     & 3.9 \\
DS-V4         & 2.7 & 3.7 & 6.2 & 3.8 & 4.1\\
GPT-5.2           & \textbf{2.1}    & 2.7    &\textbf{1.2}   & 2.7   &\textbf{ 2.2} \\
\midrule
\textbf{Average}  & \textbf{3.1} & \textbf{4.0} & \textbf{4.6} & \textbf{4.6} & \textbf{4.1} \\ 
\bottomrule
\end{tabular}
\caption{The performance difference $\Delta_\theta$ of LLMs on multi-document tasks. Results are reported on non-thinking models.}
\label{tab:qa_delta}
\end{table}
\paragraph{Impact of multilingual data}
In our setting, various models exhibit highly consistent performance on English datasets. However, in non-English contexts, the models display significant disparity. According to Figure 4, when the position of the golden document is fixed, the performance gap between most models (with the exception of DeepSeek-V3.2) remains within a narrow margin of 8 percentage points. Visually, this is reflected by the high degree of overlap among the different curves in the English plots. The curves in Figure 4 for non-English data are notably more scattered, indicating that multilingual capabilities vary significantly across different models.For additional results and detailed analysis, see Appendix~\ref{app:additional_results}.

\subsection{Results on Reasoning Models}
\begin{figure}[!tbp]
    \centering
    \begin{subfigure}[b]{0.48\textwidth}
        \centering
        \includegraphics[width=\textwidth]{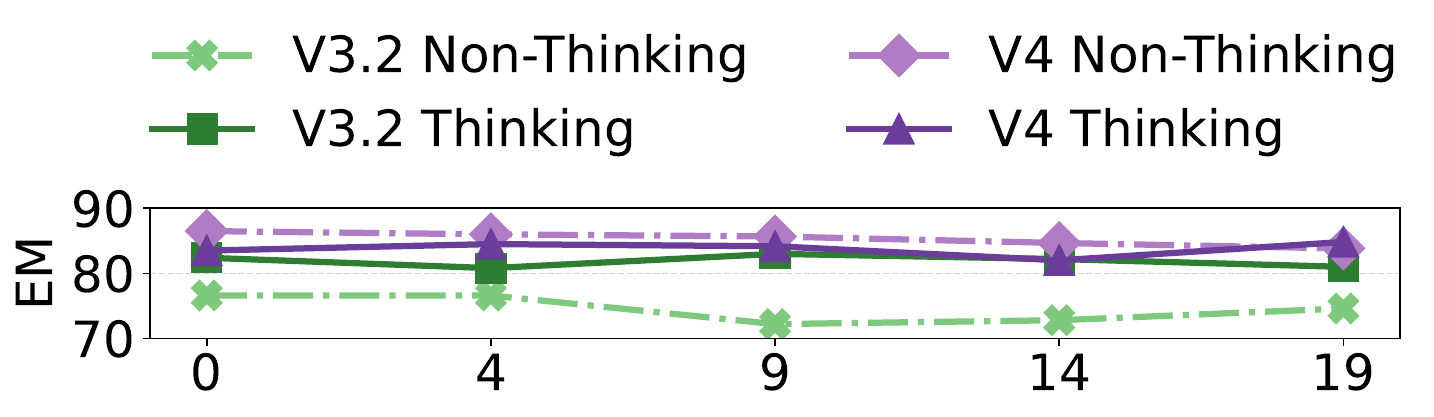} 
        \caption{English}
        \label{fig:qa_think_en}
    \end{subfigure}
    \begin{subfigure}[b]{0.48\textwidth}
        \centering
        \includegraphics[width=\textwidth]{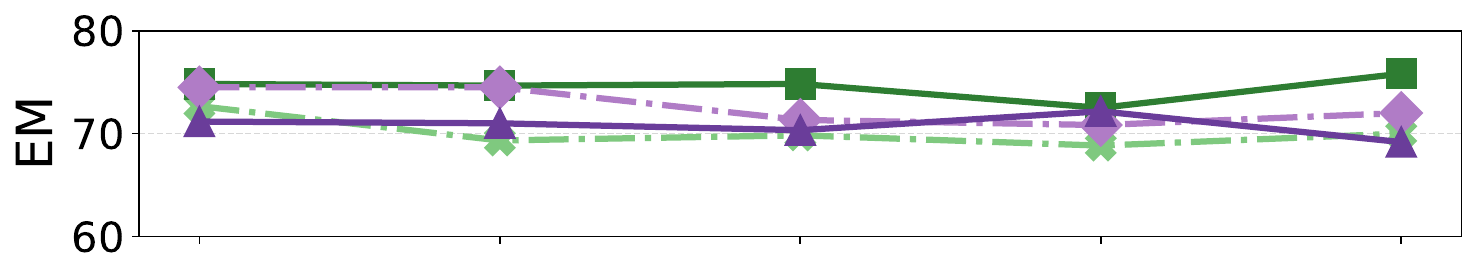}
        \caption{Chinese}
        \label{fig:qa_think_zh}
    \end{subfigure}
    \begin{subfigure}[b]{0.48\textwidth}
        \centering
        \includegraphics[width=\textwidth]{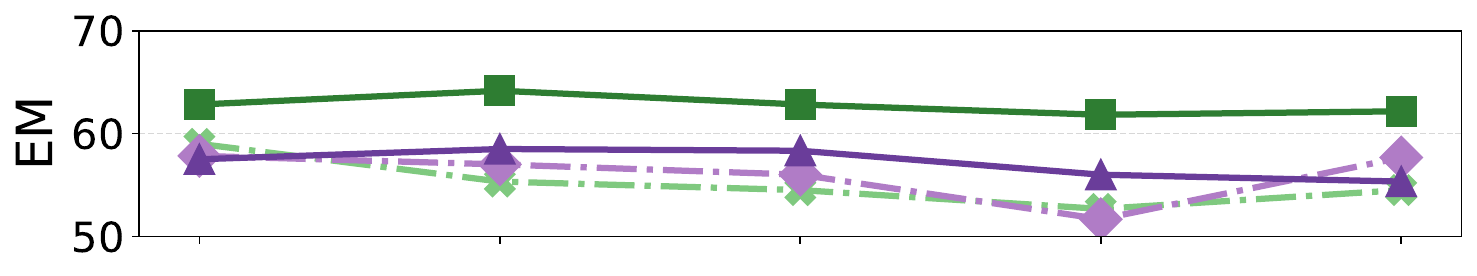}
        \caption{Arabic}
        \label{fig:qa_think_ar}
    \end{subfigure}
    \begin{subfigure}[b]{0.48\textwidth}
        \centering
        \includegraphics[width=\textwidth]{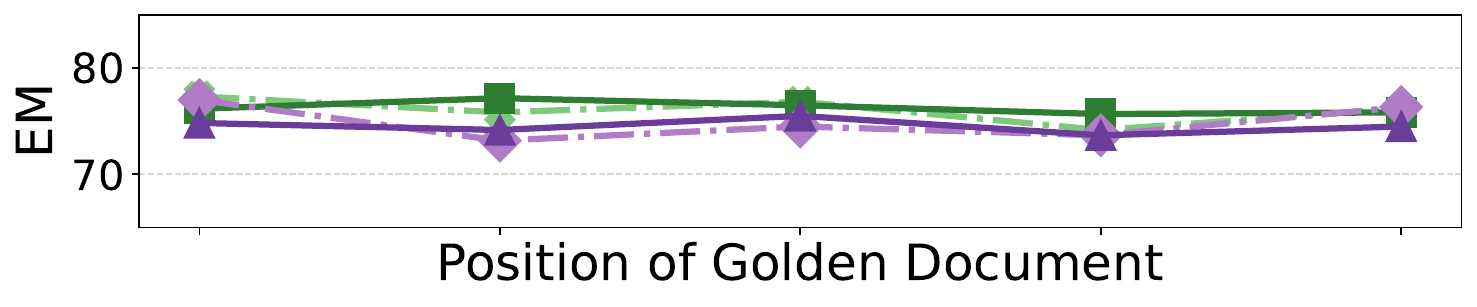}
        \caption{Spanish}
        \label{fig:qa_think_es}
    \end{subfigure}
    
    \caption{Position bias on Multi-Document tasks. We use the Deepseek-V3.2 and Deepseek-V4-Pro for evaluation.}
    \label{fig:qa_thinking}
    \vspace{-2em}
\end{figure}

Inspired by the low position bias achieved by reasoning models \cite{guo2025deepseek, liu2025deepseek, jaech2024openai}, we further investigate whether such "thinking" models can alleviate position bias in multi-document QA. As shown in Figure \ref{fig:qa_thinking}, the reasoning models exhibit more uniform performance patterns across positions compared to non-reasoning models. This observation suggests that allocating more tokens for explicit reasoning is an effective general approach to mitigating the position bias issue.
For additional analysis, see Appendix~\ref{app:thinking_analysis}.

\section{Related Work} 

The issue of \emph{position bias} was first discussed in the context of search engines, where users may be influenced by the positions of documents regardless of their content relevance \cite{joachims2017unbiased, 10.1007/978-3-030-32686-9_4}. 

Interestingly, \citet{liu2024lost} found similar behavior in large language models (LLMs). Concretely, in retrieval-augmented generation \cite{lewis2020retrieval}, the order of retrieved documents, which is permutation-invariant to humans, may significantly bias model outputs. This bias can lead to a performance drop of more than 20 points when the golden document is placed in the middle of the context for GPT-3.5 \cite{ouyang2022training}. Many efforts have been devoted to understanding this phenomenon \cite{salvatore2025lost, wu2025on, hsieh-etal-2024-found, tian-etal-2025-distance} and alleviating it \cite{wang-etal-2025-position, wang2025eliminating, he-etal-2024-never, yu-etal-2025-mitigate}.

Recent literature has shown that position bias is not only in the multi-document setting but also in multi-choice question answering. Specifically, permutations of answer options can significantly affect model performance without regard to content \cite{wei-etal-2024-unveiling, wang2025eliminating, dimino2025tracing, zhao-etal-2024-measuring, reif-schwartz-2024-beyond, brown2025order, tripathi2025pairwise}. \citet{tian2025identifying} further find that this issue also exists in multimodal language models.  

The large performance gaps across different permutations observed in previous work are counter-intuitive and motivate our study. First, multi-choice evaluation is widely used to benchmark the intelligence of LLMs. If position bias remains severe, the community may need to rethink evaluation protocols. Second, since the multi-document setting is widely adopted in many LLM applications \cite{lewis2020retrieval, hsieh2024ruler}, severe position bias can have a strong negative impact on system accuracy and trustworthiness. These concerns motivate us to conduct a comprehensive study of position bias across models, rather than mitigating the issue in limited settings, such as specific models like LLaMA-2 \cite{touvron2023llama} or GPT-3.5~\cite{ouyang2022training}.

\section{Conclusions}
This work conducts a comprehensive evaluation of modern LLMs to assess the severity of position bias in large language models. First, we confirm that small-scale LLMs, i.e., models with fewer than $32$B parameters, generally exhibit more severe position bias across all the tested languages. In contrast, scaling up model size and allowing models to reason with more tokens significantly alleviates this issue. We also find that although few-shot examples can improve overall performance, enforcing a uniform distribution of option labels in few-shot demonstrations may exacerbate position bias on LLMs less than 1T parameters. Additionally, when modern \emph{thinking} mode (i.e., generating a long reasoning chain before answering) is enabled, LLMs achieve nearly uniform performance that is largely unaffected by the position of information segments. 

We believe these findings provide many insights for the LLM community. First, evaluation protocols for small-scale models should be carefully designed, as severe answer-option bias may fail to reliably reflect model intelligence. Second, although scaling model size and test-time computation helps mitigate position bias, efficient solutions remain appealing in computation-constrained scenarios. For a detailed discussion of the limitations, please refer to Appendix~\ref{sec:limitations}. 
Ethical considerations are also addressed in Appendix~\ref{sec:ethical}.

\bibliography{custom}

\appendix

\section{Additional Results}
\label{app:additional_results}

\paragraph{The U-Shape}
As illustrated in Figure \ref{fig:qa}, we observe that the performance of most models in multi-document QA exhibits a distinct ``U-shape'' pattern. Specifically, placing the golden document at the beginning or the end of the context generally yields better performance than placing it in the middle. This observation is consistent with the findings of \citet{liu-etal-2024-lost}. However, this contrasts with our observations in multiple-choice tasks, where the first position generally achieves significantly higher performance than all others.

\section{Dataset Construction Details}
\label{app:data_construction}
We adopt the experimental setup of \citet{liu2024lost}, but since LITM only contains English data, we use the MLQA dataset for multilingual evaluation. In the LITM dataset, each question is associated with $K$ independent documents, only one of which (the "golden documeLITMnt") contains the answer. We conduct our experiments using a configuration of $N=20$ documents, placing the golden document at positions 0, 4, 9, 14, and 19.

Because MLQA provides three-way alignment among \textless question, answer, golden doc\textgreater, we randomly sample $N-1$ irrelevant documents from the same dataset for each question to simulate the setup in \citet{liu2024lost}. Following the LITM configuration, we construct datasets for English, Arabic, Chinese, and Spanish with $N=20$ and golden documents at positions 0, 4, 9, 14, and 19.
For each language, we randomly sample 500 data points consisting of \textless question, answer, docs\textgreater\ and evaluate model performance at each of the five golden document positions. Consequently, each experimental run has $5 \times 500 = 2,500$ examples.

\section{Additional Analysis of Thinking Models}
\label{app:thinking_analysis}

We further analyzed why thinking models (long CoT) outperform non-thinking models in the multi-document scenario. To this end, we utilized the thinking and non-thinking models of the DeepSeek-V3.2 for a controlled comparison. We categorize the cases where thinking models exhibit an advantage into three types of errors made by the non-thinking baseline, which the thinking models successfully resolve.

\begin{table}[h]
\centering
\begin{tabular}{lc} 
\toprule
\textbf{Error Type} & \textbf{Percentage} \\
\midrule
Type 1: Conflicting Docs & 38.4\% \\
Type 2: Relevance Identification & 27.7\% \\
Type 3: Unknown & 33.9\% \\
\bottomrule
\end{tabular}
\caption{Error type distribution among cases where thinking models outperform non-thinking models.}
\label{tab:error_types}
\end{table}
We find that in 66.1\% of cases (Type 1 and Type 2 combined), the thinking model's improved performance is attributed to its ability to comprehensively summarize multiple relevant documents, analyze and compare conflicting information, and better identify relevant documents.

\subsection{Definition of Error Types}
We define the three error types as follows:

\textbf{Type 1 — Conflicting Docs.} The non-thinking model fails to compare or reconcile conflicting information presented in the context.

\textbf{Type 2 — Relevance Identification.} Although model outputs frequently cite supporting documents, the non-thinking model does not include the most relevant one among them.

\textbf{Type 3 — Unknown.} There are no explicit clues in the model-generated answer or the reference answer to explain why the thinking model performs better.

\subsection{Data Selection Criteria}

For each question, with the golden document placed at positions 0, 4, 9, 14, and 19, five completely incorrect responses received a score of 0. Each additional correct response earned +1 point, yielding a maximum score of 5. We identified 65 questions where the thinking model outperformed the non-thinking model.

\section{Limitations}
\label{sec:limitations}
Our study is primarily limited by the scope of language and prompt variety. While we evaluate five diverse languages, the results may not generalize to extremely low-resource languages with distinct syntactic structures.  Finally, while thinking effectively mitigate bias, they do so at the cost of significantly increased inference-time computation and latency, which may be impractical for real-time applications.

\section{Ethical Considerations}
\label{sec:ethical}
The primary ethical concern of this work involves the environmental impact and computational cost of evaluating extremely large-scale models across multiple permutations and reasoning modes. To address this, we utilized down-sampled datasets where consistent trends were observed. Additionally, our findings highlight that position bias can lead to misleading performance; failing to account for this bias could result in the deployment of unreliable systems in high-stakes domains. We encourage the community to use our findings to develop more robust and fair evaluation protocols rather than to manipulate model performance through strategic information placement.

\end{document}